\pdfoutput=1

\documentclass[11pt]{article}

\input{premable.sty}

\title{
Iterative Refinement of Project-Level Code Context for Precise Code Generation with Compiler Feedback
}

\author{
Zhangqian Bi$^1$\footnotemark[1]
~~~~ Yao Wan$^1$\footnotemark[1]
\footnotemark[2]
~~~~ Zheng Wang$^2$~~~~ Hongyu Zhang$^3$ ~~~~ Batu Guan$^1$  \\
  \textbf{Fangxin Lu}$^1$~~~~ \textbf{Zili Zhang}$^4$~~~~ \textbf{Yulei Sui}$^5$~~~~ 
  \textbf{Hai Jin}$^1$\footnotemark[1]~~~~ 
  \textbf{Xuanhua Shi}$^1$\footnotemark[1]\\ 
  $^1$Huazhong University of Science and Technology ~~~~ $^2$University of Leeds\\
  $^3$Chongqing University ~~~~ 
  $^4$Shanghai Jiao Tong University ~~~~ $^5$University of New South Wales \\
  \texttt{\{zqbi,wanyao,hjin,xhshi\}@hust.edu.cn}
}

\begin{document}
\maketitle
\renewcommand{\thefootnote}{\fnsymbol{footnote}}
\footnotetext[1]{Also with National Engineering Research Center for Big Data Technology and System, Services Computing Technology and System Lab, Cluster and Grid Computing Lab, School of Computer Science and Technology, Huazhong University of Science and Technology, Wuhan, 430074, China.}
\footnotetext[2]{Yao Wan is the corresponding author.}

\begin{abstract}
\textit{Large Language Models} (LLMs) have shown remarkable progress in automated code generation. Yet, LLM-generated code may contain errors in API usage, class, data structure, or missing project-specific information.
As much of this project-specific context cannot fit into the prompts of LLMs, we must find ways to allow the model to explore the project-level code context. 
We present \tool, a new code generation approach that uses compiler feedback to improve the LLM-generated code.
\tool first leverages static analysis to identify 
mismatches between the generated code and the project's context.
It then iteratively aligns and fixes the identified errors using information extracted from the code repository.
We integrate \tool with two representative LLMs, i.e., GPT-3.5-Turbo and Code Llama (13B), and apply it to Python code generation.
Experimental results show that \tool significantly improves the vanilla LLMs by over 80\% in generating code dependent on the project context and consistently outperforms the existing retrieval-based code generation baselines.
\end{abstract}

\section{Introduction}
\textit{Large Language Models} (LLMs), especially those pre-trained on code,
as demonstrated by tools such as GitHub Copilot \cite{copilot}, Amazon's CodeWhisperer \cite{amazon2023ai}, and ChatGPT~\cite{chatgpt}, 
are revolutionizing how developers approach programming by automatically generating code for given contexts (\eg, natural-language descriptions or surrounding incomplete code). 
While existing LLM-based code generation tools excel in code generation within a small-scale and isolated context (\ie, a single file), integrating LLM-based code generation into real-world software projects remains challenging~\cite{li2024deveval}.

Practical code generation for software repositories is often associated with broader project-level contexts declared in other repository files due to the demands of 
modularity, structure, and comprehension in software management~\cite{kemerer1995software}.
Solely providing the task requirements and surrounding incomplete code can lead LLMs to overlook the complex hierarchies of APIs, classes, data structures, or type constraints specific to a software project's repository, risking omitting essential logic during code generation~\cite{gifany2013logic}.


\begin{figure}[t]
\begin{center}
\centerline{\includegraphics[width=1\columnwidth]{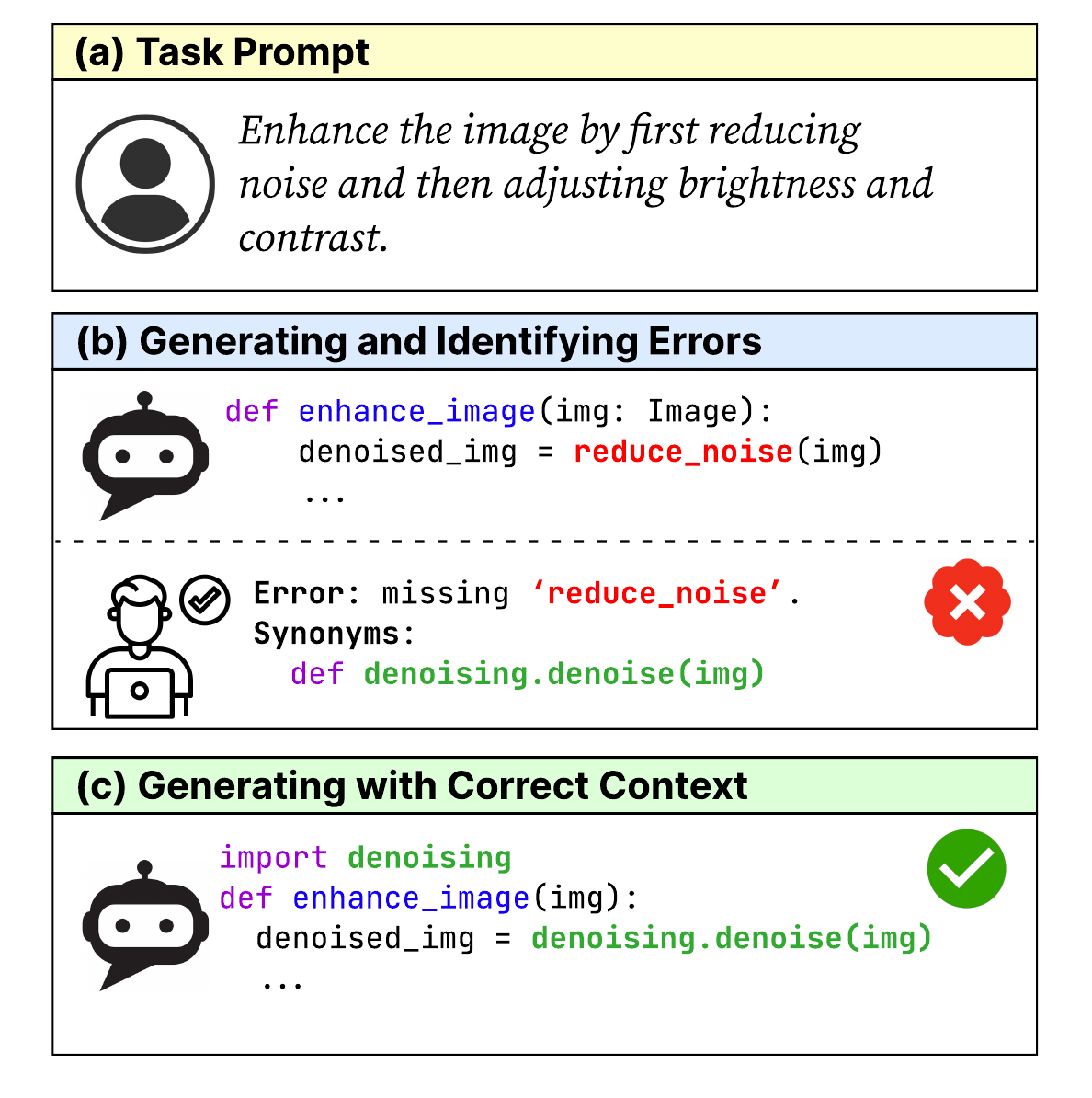}}
\vspace{-1em}
\caption{
LLM-based code generation example. (\textbf a) task prompt; (\textbf b) wrong solution and error identification; (\textbf c) correct solution utilizing project context
}
\label{fig:illustrative_example}
\end{center}
\vspace{-3em}
\end{figure}

As a motivating example, consider the case shown in Figure~\ref{fig:illustrative_example}. 
In this case, we use the OpenAI GPT-3.5-Turbo API to generate an image-enhancing function by first reducing the notice and then adjusting brightness and contrast.
Given the prompt in Figure~\ref{fig:illustrative_example}a, the LLM will produce a code snippet of calling a \texttt{reduce\_noise} method as depicted in Figure~\ref{fig:illustrative_example}b.
Although the generated code follows a standard workflow specified in the task prompt, it leads to a compilation error in our application context because \texttt{reduce\_noise} is not implemented.
This is unsurprising as the prompt does not provide enough project-level context for the LLM, and can be fixed by providing the project-specific function \texttt{denoising.denoise} as context, as illustrated in Figure~\ref{fig:illustrative_example}c.
While a carefully engineered prompt may resolve this issue, it is not always possible for the user to generate such prompts, and the project context may be too large to fit into the prompt. As we will empirically show later in (Section~\ref{sec:error_analysis}), such errors are commonly found when applying LLMs to repository-level code generation~\cite{li2024deveval}.   

This paper investigates a way to effectively integrate LLM-based code generation with existing code implementations within a software project. Our solution is to leverage project-level contextual information, such as project-specific implementation of classes, methods, and data structures, to reduce compilation errors and improve code quality. Directly incorporating the entire project code into a language model is infeasible due to model input sequence length limitations. Instead, we use compiler-based analysis to post-process the model-generated code by first detecting discrepancies between the generated code and the project's context. We then utilize information extracted from the project code base to rectify the mismatches in using modules, APIs, and classes. Our approach combines well-established compiler techniques with emerging generative methods, allowing software developers to leverage the power of LLMs without being overwhelmed and discouraged by the frequent compilation and semantic errors in the model-generated code.

We present \tool, a method to allow an LLM to leverage the code repository of a software project to enhance the quality of the generated code. 
For a given LLM-generated code sample, 
 \tool first compiles it and identifies context-related errors.
It then retrieves the related context from the code repository to fix the errors.
This iterative generation and verification process proceeds repetitively until no error is identified in the generated solution.
We demonstrate that \tool enhances the accuracy of generation (as indicated by pass rates) by bridging the gap between the repository context and the intended solution.

We evaluate \tool by applying it to the CoderEval benchmarking dataset \cite{yu2023codereval}, which consists of code generation tasks utilizing project-specific context.
We test \tool on two popular code generation models: the GPT-3.5-Turbo~\cite{chatgpt} and Code Llama~\cite{roziere2023code}.
Experimental results demonstrate that \tool significantly improves the repository-level code generation performance of different dependency levels, outperforming the baseline by over 80\% relative pass rates in generating functions dependent on project-specific contexts. 
Moreover, our iterative method consistently enhances the performance of vanilla retrieval-augmented generation.
We also provide a comprehensive analysis of the effectiveness and limitations of \tool, offering insights for future research.

This paper makes the following contributions:
\begin{itemize}[left=0pt,itemsep=0pt]
    \item An empirical study to analyze the error distribution in self-contained and repository-level code generation, highlighting the significance of precise and grounded program context in generating code at the project level (Section \ref{sec:error_analysis});
    \item A new iterative generation-verification-retrieval method that leverages the program compiler to eliminate context-related errors in repository-level code generation (Section \ref{methodology});
    \item Extensive experiments and analysis based on two LLMs, i.e., GPT-3.5-Turbo and Code Llama (13B), showing the effectiveness of the proposed \tool method (Section~\ref{sec_experiment_results}).
\end{itemize}
The source code and dataset used in this paper are available at: \url{https://github.com/CGCL-codes/naturalcc/tree/main/examples/cocogen}.

\section{Background}

\subsection{LLM-based Code Generation}

Our work targets the code generation task, which produces source code from a natural-language description complemented by programming context (\textit{e.g.}, project-specific APIs, and data structures). 
We denote this input as $x$.
Given $x$, it is first converted to a sequence of tokens $\bm x = [x_1, \ldots, x_{|\bm x|}]$, and a generative \textit{Language Model} (LM) $p_{\text{LM}}(\bm x)$ predicts new tokens sequentially. 
At each step $t$, the LM calculates the probability distribution of the next token as $p_{\text{LM}}(x_t | x_{1:t-1})$.
The probability of generating a program $y$ with token sequence $\bm y = [x_{|\bm x|+1}, \ldots, x_{|\bm x|+|\bm y|}]$ is computed as a product of next-token distributions given left context:
\begin{equation}
p(y|x) = \prod_{t=|\bm x|+1}^{|\bm x| + |\bm y|}p_{\text{LM}}(x_t|x_{1:t})
\end{equation}
\indent For few-shot learning with large LMs, the generation is also often conditioned on a fixed set of $m$ exemplars, $\{\langle x_i, y_i\rangle\}_{i \le m}$.
Thus, the LLM-based code generation can be formulated as:
\begin{equation}
    p_{\text{LM}}(y|x) = p(y | x, \{\langle x_i, y_i\rangle\}_{i \le m})
\end{equation}
\indent Practically, the probability of the next token $x_t$ depends on a fixed number of preceding tokens $x_{\max(1, t-w)}:x_{t-1}$, defined by the model's context window length $w$, without encompassing the entire software project's code base.


\subsection{Error Analysis in Code Generation}
\label{sec:error_empirical}
\begin{table}[t]
\caption{
Typical errors reported in compilation and execution
}
\label{table:error_examples}
\vspace{-1em}
\begin{center}
\resizebox{0.5\textwidth}{!}{
\begin{tabular}{p{1.9cm}|l}
\hline
Error Type& Example                                                               \\ \hline
\textit{UNDEF}      & No name `AsyncBolt5x0' in module 'neo4j.\_sync.io.\_bolt5'            \\
\textit{API}        & No value for argument `xmls' in function call                         \\
\textit{OBJECT}     & `function' object is not subscriptable                                \\
\textit{FUNC}       & The generated function not passes a test case                         \\
\textit{OTHER}      & Parsing failed: `expected an indented block after function definition' \\ \hline
\end{tabular}

}
\end{center}
\vspace{-1em}
\end{table}

\label{sec:error_analysis}
The performance of simple function-level code generation has significantly improved, as demonstrated by an increase in the pass rate from 31.6\% with CodeT5~\cite{wang2021codet5} to 94.7\% with the state-of-the-art Code Llama~\cite{roziere2023code} on the widely-used HumanEval benchmark~\cite{chen2021evaluating}.
However, a recent study~\cite{yu2023codereval} shows that existing LLMs for code generation struggle to generate code snippets that are dependent on the project contexts, such as private APIs, classes, data structures, or type constraints.
To this end, 
various benchmarks, including ClassEval~\cite{du2023classeval}, CoderEval~\cite{yu2023codereval}, and CrossCodeEval~\cite{ding2024crosscodeeval}, have been devised to assess the performance of LLMs in generating context-dependent code within the project.

To better illustrate our motivation, we perform an empirical analysis of when the LLMs fail to generate complex code that is dependent on project-level context, on the CoderEval dataset~\cite{yu2023codereval}.
This dataset comprises 85 function-level tasks and 145 project-level tasks.
Specifically, we select the GPT-3.5-Turbo~\cite{chatgpt} 
as a target LLM for function-level code generation.
Based on it, we select a state-of-the-art method called RepoCoder~\cite{zhang2023repocoder}. RepoCoder retrieves project-level context as an augmentation and incorporates the five code fragments with the highest similarity scores, as determined by dense passage retrieval~\cite{karpukhin2020dense}, into the prompt for improved code generation.

\begin{figure}[t]
    \centering
    \includegraphics[width=1\columnwidth]{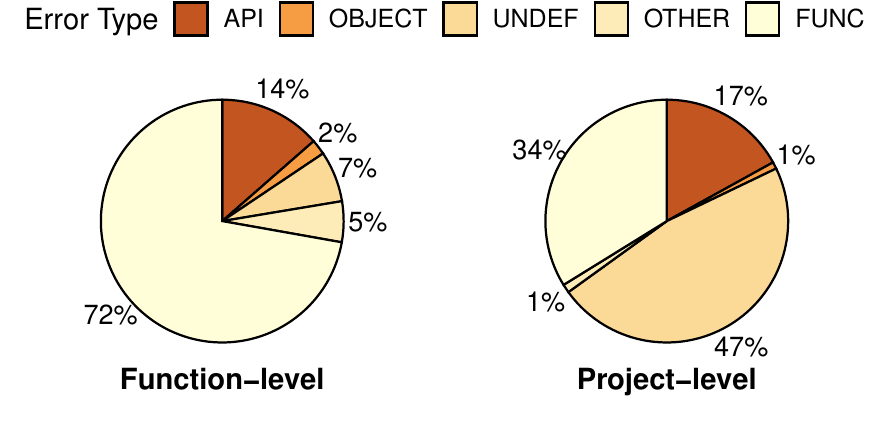}
    \vspace{-1em}
    \caption{Distribution of error types in the generated solutions on CoderEval dataset
    }
    \label{fig:error_types}
    \vspace{-1em}
\end{figure}

\begin{figure*}[t]
    \centering
    \includegraphics[width=1.0\linewidth]{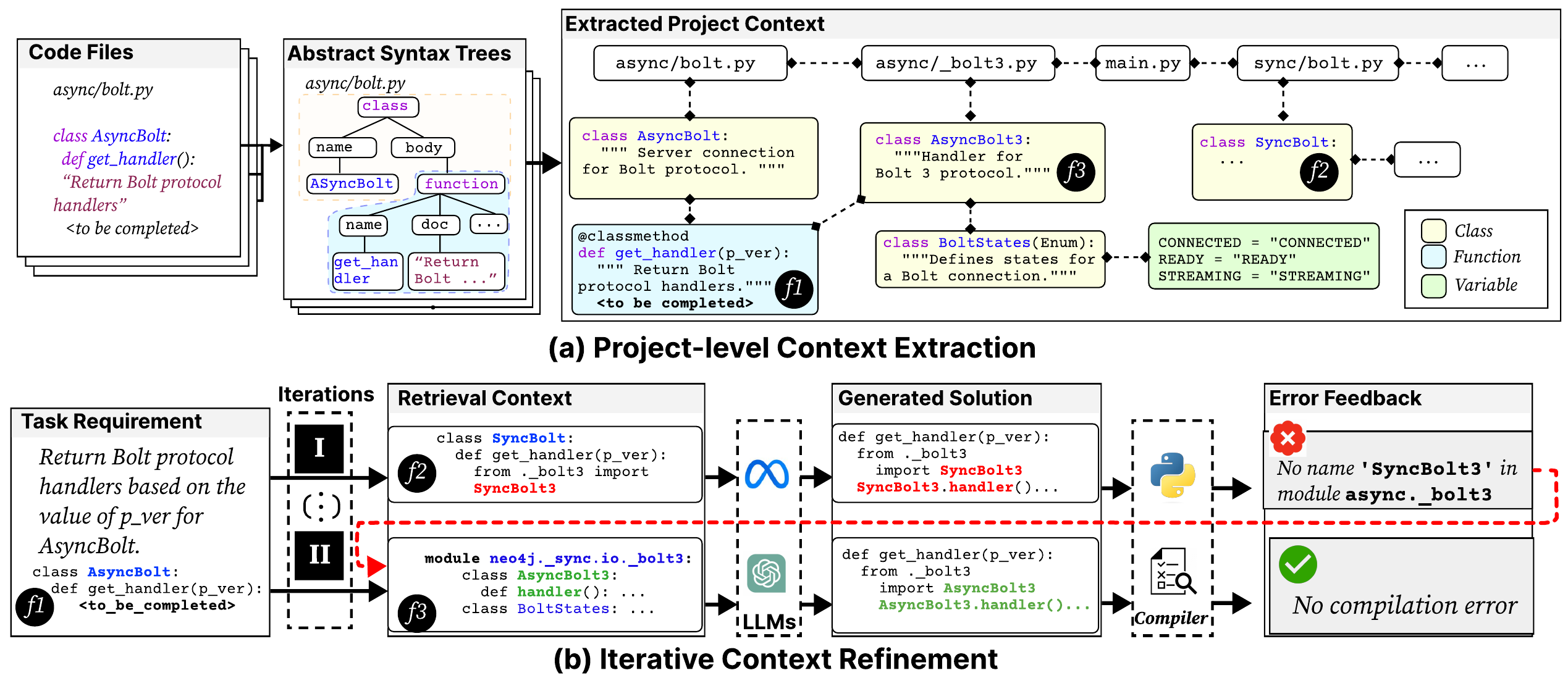}
    \vspace{-1.2em}
    \caption{
    Overview of the \tool method. (\textbf a) the project-level code context extraction process; (\textbf b) iterative refinement to fix compiler-reported errors
}
\vspace{-1em}
    \label{fig:overview}
\end{figure*}

As the code is generated, we compile it and collect any errors reported by the compiler or encountered during testing.
We have generated 10 candidate solutions for a task, comprising 850 solutions for function-level tasks and 1450 for project-level tasks. We report the Pass@10 rate as 53.57\% for function-level tasks and 39.73\% for project-level tasks. The code that does not pass the test has been taken into error analysis.
Each generated code snippet contains precisely one type of error. If multiple errors are reported by the compiler, the most common error type is selected for analysis.
The error distribution reveals that four specific types of errors constitute the majority of all errors encountered. We categorize these errors into:
1) \textit{UNDEF}, involving Use of Undefined Symbol, 2) \textit{API}, involving Incorrect Use of APIs, 3) \textit{OBJECT}, involving Improper Use of an Object, 4) \textit{FUNC}, involving Runtime or Functional Errors, and 5) \textit{OTHER}, involving Other Syntax and Semantic Errors.
Table~\ref{table:error_examples} presents several
errors encountered in compiling and testing.
One example is the ``\textit{UNDEF}'' error, where a variable \texttt{AsyncBolt5x0} is referenced but does not exist in the specified module.

Figure~\ref{fig:error_types} illustrates the distribution of error types, under the scenario of function-level code generation and project-level code generation.
From this figure, we can observe that the majority of errors are runtime or functional errors, accounting for $72\%$ and $34\%$ for function-level and project-level code generation, respectively.
Furthermore, the \textit{UNDEF} errors and the \textit{API} errors account for substantially high portions of $21\%$ and $64\%$ for function-level and project-level code generation, respectively.
\tool addresses both types of errors associated with project context by supplying the relevant project context. 
Experiments demonstrate that \tool not only fixes these two types of error but also mitigates other compilation errors by providing context feedback on error messages to the code LM, leading directly to an improvement in prediction accuracy.

\section{Methodology}
\label{methodology}
\subsection{Overview}
Figure~\ref{fig:overview} depicts the workflow of \tool, consisting of two crucial components: 1) a method for extracting project-level code context through both syntactic and semantic approaches, and 2) a component responsible for iterative generation and evaluation of solutions. This process refines the generated solutions incrementally, ensuring they evolve towards an error-free state that seamlessly aligns with the codebase of the software project.


\subsection{Project-Level Code Context Extraction}

Supposing that the code generation tools are activated at a specific juncture. 
In light of the natural-language requirement and the code produced by LLMs after an initial iteration, our objective is to
extract the semantic context of the generated code from the project's code base.

Unlike plain texts, source code has syntactic structures that enable precise identification of elements in a project.
Thus, in practice, we employ syntax-directed program analysis~\cite{alfred2007compilers} at various points throughout the offline stage to extract the code context at the project level.
We initially employ a parser to transform each source code file within the project into an \textit{Abstract Syntax Tree} (AST), extracting tree nodes that correspond to classes, functions, or variables. 
Subsequently, if a node of these types is found to be a child of another node
(\textit{e.g.}, the function \texttt{get\_handler} and the class \texttt{AsyncBolt} in Figure~\ref{fig:overview}a's AST)
, an edge is created from the parent node to the child node, establishing a hierarchical relationship.

Take the function 
$f_1$
from Figure~\ref{fig:overview}a as an example.
From this figure, we can see that both its semantics (\textit{e.g.}, its docstring), and its syntactic relation between the parent class \texttt{AsyncBolt} and file \texttt{async/bolt.py} are captured.
This allows \tool to find the function from the project syntactically using the function's qualified name \texttt{AsyncBolt.get\_handler()}, or semantically according to its docstring \textit{``Return Bolt protocol handlers''}.

\subsection{Retrieval-Augmented Code Generation}
\label{sec:retrieval_augmented}
We leverage project-level code context in the retrieval-augmented generation paradigm~\cite{zhang2023repocoder,ding2022cocomic,karpukhin2020dense}, which has been widely adopted to integrate factual knowledge into LLMs and address hallucination issues.
In practice, we commence by extracting project-level context from the database through the construction of a \textit{Structured Query Language} (SQL) query. Following this, we enhance the acquired context by retrieving similar code snippets based on the dense passage retrieval techniques~\cite{karpukhin2020dense}.

\begin{figure}
    \centering
    \includegraphics[width=\linewidth]{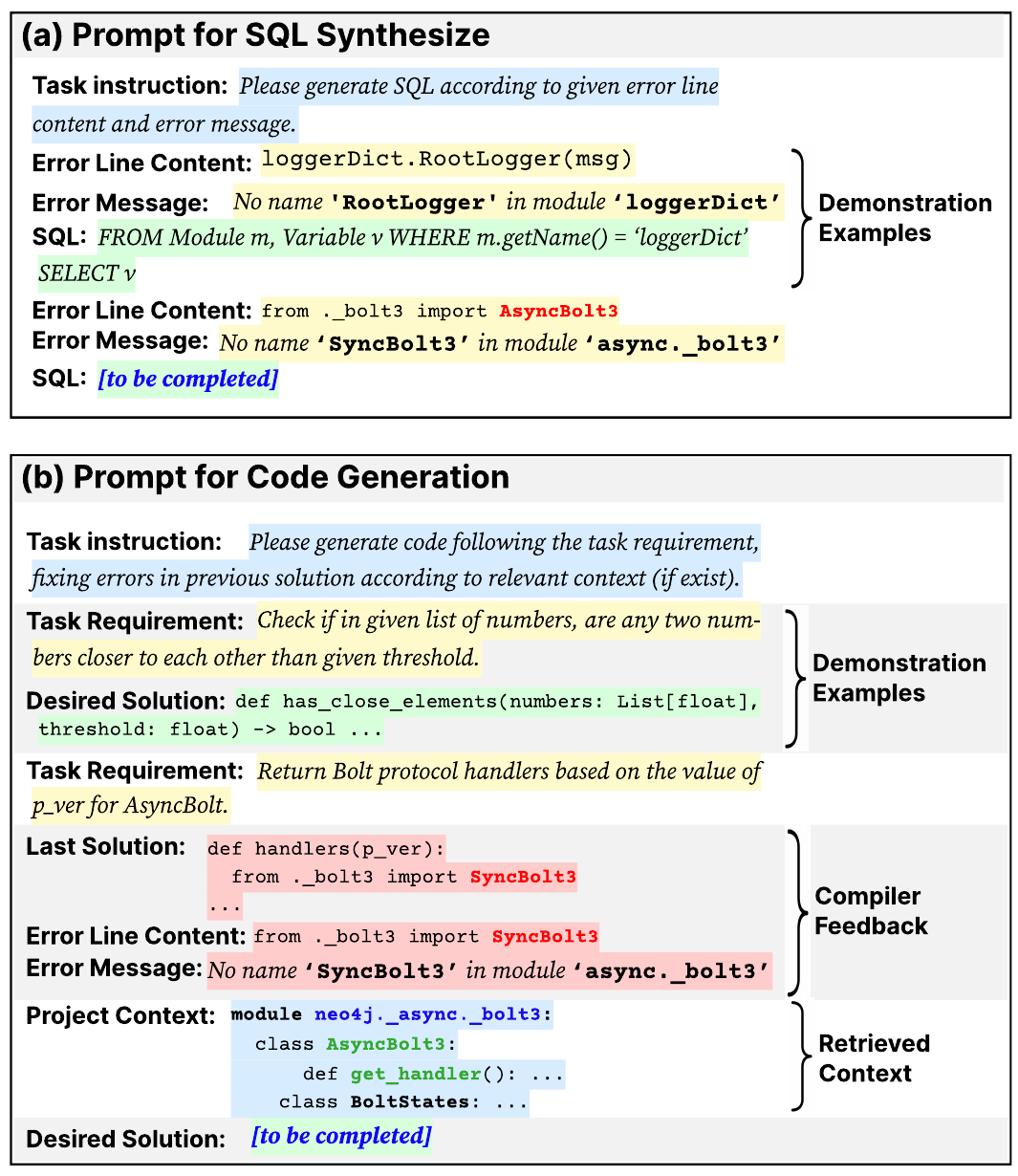}
    \vspace{-1em}
    \caption{Prompt examples for (\textbf a) SQL synthesize and (\textbf b) code generation}
    \vspace{-1em}
    \label{fig:prompt_examples}
\end{figure}

\paragraph{Structural Search.}
Based on the compiler feedback, we aim to retrieve the relevant project-level context from all extracted ones.
We implement this by transforming the textual compiler feedback into an SQL query using the ChatGPT.
The prompt used is presented in Figure~\ref{fig:prompt_examples}a.
Several examples of paired compiler feedback and SQL queries are provided as demonstrations for in-context learning\footnote{Only one demonstration example is illustrated in Figure~\ref{fig:overview}(b), the full list of examples can be found in Appendix~\ref{appendix:example_queries}.}.

For instance, consider the compiler feedback:
\textit{No name \texttt{'SyncBolt3'} found in module \texttt{'async.\_bolt3'}},
the resulting SQL query generated by ChatGPT is as follows:

\definecolor{keywordcolor}{HTML}{9C00D1}
\definecolor{stringliteralcolor}{HTML}{8B2252}
\definecolor{variablecolor}{HTML}{1400F9}

\begin{tcolorbox}[colback=white, colframe=black, sharp corners, left=1pt,right=1pt, top=1pt, bottom=1pt,boxrule=1pt]
\textcolor{keywordcolor}{\textbf{\texttt{FROM}}}
\textit{Module \textcolor{variablecolor}{m}, Class \textcolor{variablecolor}{c}} \\
\textcolor{keywordcolor}{\textbf{\texttt{WHERE}}}
\textit{\textcolor{variablecolor}{m}.contains(\textcolor{variablecolor}{c})} \\
  \textcolor{keywordcolor}{\textbf{\texttt{and}}}
\textit{\textcolor{variablecolor}{m}.getName() = \textcolor{stringliteralcolor}{`async.\_bolt3'} } \\
\textcolor{keywordcolor}{\textbf{\texttt{SELECT}}}
\textit{\textcolor{variablecolor}{m}, \textcolor{variablecolor}{c}}
\end{tcolorbox}
Using this SQL query, the code snippets that involve the implementations of \texttt{AsyncBolt3} will be returned from our constructed database.
The detailed process of constructing and querying such SQL database is refer to Appendix~\ref{apdx:sql_system}.


\paragraph{Semantic Search.}
In addition to returning the project-level context retrieved by the SQL query, we also enhance the acquired context by retrieving similar code snippets
using dense passage retrieval~\cite{karpukhin2020dense}.
In the initial search round, no compilation error is reported, and \tool utilizes the task description string for retrieval. In subsequent searches, \tool employs the error report and the corresponding error line.

%

Given a natural-language query $q$, \tool first converts it into an embedding vector by utilizing an encoder network, as follows:
\begin{equation}
    \mathbf{h}_{q} = \text{ENCODER}(q)
\end{equation}
\indent \tool utilizes a pre-trained Transformer  network~\cite{vaswani2017attention} as the encoder.
After generating the query vector, \tool calculates the cosine similarity between the query and embedding vectors of each context entry $\bm h_{c}$. This similarity measure is defined as:
\begin{equation}
    sim(\mathbf{h}_{q}, \mathbf{h}_{c}) = \dfrac{\mathbf{h}_{q}^\intercal \mathbf{h}_{c}}{||\mathbf{h}_{q}||\cdot ||\mathbf{h}_{c}||}
\end{equation}
and the top-$n$ entries exhibiting the highest similarity to the query are retrieved as results.

\begin{table*}[t]
\caption{
Pass rates of \tool based on two LLMs, i.e., GPT-3.5-Turbo and Code Llama (13B), assessed against various baselines across different splits of the CoderEval dataset
}
\label{table:main_result}
\vspace{0em}
\begin{center}
\resizebox{1\textwidth}{!}{
\begin{tabular}{lrrrlrrrlrrr}
\hline
Data Split & \multicolumn{3}{c}{Class Runnable}               &  & \multicolumn{3}{c}{File Runnable}                                  &  & \multicolumn{3}{c}{Project Runnable}                               \\ \cline{1-4} \cline{6-8} \cline{10-12} 
Method     & Pass@1         & Pass@5         & Pass@10        &  & Pass@1               & Pass@5               & Pass@10              &  & Pass@1               & Pass@5               & Pass@10              \\ \hline
\multicolumn{4}{l}{\textit{LLM: GPT-3.5-Turbo}}                        &  & \multicolumn{1}{l}{} & \multicolumn{1}{l}{} & \multicolumn{1}{l}{} &  & \multicolumn{1}{l}{} & \multicolumn{1}{l}{} & \multicolumn{1}{l}{} \\
\rowcolor[HTML]{EFEFEF} 
Direct    & 8.73           & 12.57          & 14.55          &  & 19.85                & 27.62                & 30.88                &  & 9.57                 & 12.08                & 13.04                \\
ReACC      & 20.36          & 33.27          & 38.18          &  & 17.65                & 28.92                & 33.82                &  & 11.30                & 19.53                & 21.74                \\
\rowcolor[HTML]{EFEFEF} 
RepoCoder  & \textbf{35.45} & 40.46          & 41.82          &  & 29.41                & 34.61                & 36.76                &  & 16.96                & 19.57                & 21.74                \\
\tool      & 28.00          & \textbf{44.92} & \textbf{49.09} &  & \textbf{30.29}       & \textbf{43.58}       & \textbf{47.06}       &  & \textbf{21.30}       & \textbf{36.73}       & \textbf{39.13}       \\ \hline
\multicolumn{4}{l}{\textit{LLM: Code Llama (13B)}}                     &  & \multicolumn{1}{l}{} & \multicolumn{1}{l}{} & \multicolumn{1}{l}{} &  & \multicolumn{1}{l}{} & \multicolumn{1}{l}{} & \multicolumn{1}{l}{} \\
\rowcolor[HTML]{EFEFEF} 
Direct    & 18.91          & 30.65          & 34.55          &  & 18.53                & 27.82                & 29.41                &  & 5.22                 & 8.70                 & 13.04                \\
ReACC      & 20.36          & 33.27          & 38.18          &  & \textbf{17.65}       & 27.61                & 33.82                &  & 11.30                & 19.53                & 21.74                \\
\rowcolor[HTML]{EFEFEF} 
RepoCoder  & 17.82          & 35.22          & 40.00          &  & 15.00                & 28.31                & 32.35                &  & \textbf{16.09}       & 21.36                & 21.74                \\
\tool      & \textbf{26.36} & \textbf{39.42} & \textbf{41.82} &  & 17.06                & \textbf{29.39}       & \textbf{33.82}       &  & 13.04                & \textbf{28.04}       & \textbf{34.78}       \\ \hline
\end{tabular}

}
\end{center}
\vspace{1em}
\end{table*}

\subsection{Refinement with Compiler Feedback}
Figure~\ref{fig:overview}(b) showcases the iterative refinement pipeline. 
Given the task requirement and partial function 
$f_1$,
a semantic retrieval is activated to identify similar functions.
Specifically, the function $f_2$,
which provides equivalent functionality in synchronous scenario, is identified.
Utilizing both the retrieved context and the prompt illustrated in Figure~\ref{fig:prompt_examples}(b), the language model generates a solution.

However, the generated output mistakenly invokes \texttt{SyncBolt3} due to its intended use in asynchronous scenarios, not aligning with the synchronous scenario in $f_2$
The compiler's feedback highlights this error.
With this feedback, \tool conducts the structural and semantic search, leading to the discovery of the correct function 
$f_3$
for asynchronous scenarios.
Incorporating the error details and context into the next iteration ensures accurate function invocation.
This process goes iterative until no error is reported by the compiler, resulting in an error-free solution that aligns with the project's environment.

It is noteworthy that \tool does not take into account FUNC errors, which arise during execution despite successful compilation. 
\tool focuses on addressing compilation errors, which constitute 66\% of the total errors in the context of project-level code generation, as shown in Figure~\ref{fig:error_statistic}.

\section{Experimental Setup}

\subsection{Models and Datasets}
To validate the effectiveness of \tool,
we select GPT-3.5-Turbo and Code-Llama 13B base\footnote{\url{https://huggingface.co/codellama/CodeLlama-13b-hf}} language models for investigation. The technical details of models and their invocation for inference are presented in Appendix~\ref{apdx:models}.

We conduct experiments using the Python split of the CoderEval benchmark~\cite{yu2023codereval}, referred to as CoderEval-Python. 
It is a benchmark designed to evaluate models within realistic software development scenarios. Without loss of generalizability, we concentrate on the Python programming language within this dataset.
This benchmark categorizes 230 test samples into six levels of context dependency:
1) \textit{self-contained}: built-in types/functions, no imports required;
2) \textit{slib-runnable}: standard libraries/modules, no installation needed;
3) \textit{plib-runnable}: publicly-available libraries on PyPI/Maven;
4) \textit{class-runnable}: code outside the function but within a class;
5) \textit{file-runnable}: code outside the class but within the file;
and
6) \textit{project-runnable}: code in other source files.
We concentrate on the last three dependency types, where the solutions are dependent on project-specific contexts.
There are 55, 68, and 23 tasks associated with each dependency level, respectively.

We also evaluate \tool on two function-level code generation benchmarks, namely HumanEval~\cite{chen2021evaluating} and MBPP~\cite{austin2021program}, as well as a project-level code completion benchmark named CrossCodeEval~\cite{ding2024crosscodeeval}, to further validate its generalizability across other coding tasks.
the statistics and experimental results of these benchmarks can be found in Appendix~\ref{apdx:dataset} and Appendix~\ref{apdx:more_experiments}.

\subsection{Baseline Methods}
\tool can function seamlessly and be integrated into existing LLMs, requiring only black-box access to these models.
In this paper, we select two state-of-the-art LLMs for code generation, namely GPT-3.5-Turbo~\cite{chatgpt} and Code Llama (13B)~\cite{roziere2023code}, as our base models.
To validate the effectiveness of \tool, we compare it with the following baselines:

    \noindent$\triangleright$ \textbf{Direct} Generation~\cite{yu2023codereval}. 
    This line of method denotes directly inputting the task requirements into LLM for code generation, without providing additional context.
    
    \noindent$\triangleright$ \textbf{ReACC}~\cite{lu2022reacc}. 
    We employ the retrieval-augmented generation technique introduced in this baseline for code generation tasks. More precisely, we retrieve project contexts aligned with the task instructions 
 semantically through embedding similarity, and leverage them to augment the prompts of LLMs for better code generation.
    
    \noindent$\triangleright$ \textbf{RepoCoder}~\cite{zhang2023repocoder}. 
    Similar to our work, the referenced baseline also proposes the iterative refinement of generated code. Specifically, it involves retrieving similar code snippets derived from the previously generated ones, and employing them to augment the prompts of LLMs. One distinguishing feature is that this baseline does not leverage compiler feedback.

\subsection{Evaluation Metrics}
We employ the Pass@$k$ metric~\cite{chen2021evaluating, yu2023codereval} for code generation tasks, and employ the code exact match (C-EM), code edit similarity (C-ES), identifier exact match (I-EM), and identifier F1 score (I-F1) to evaluate both the code match rate and the identifier match rate for code completion tasks, follows~\citet{ding2024crosscodeeval}.
We present the details of the Pass@$k$ metric for code generation. For metrics used in the code completion task, readers are referred to Appendix~\ref{apdx:evaluation_metrics}.

\paragraph{Pass@$k$.} Following previous studies~\cite{chen2021evaluating, yu2023codereval}, we evaluate the functional correctness of the generated code by executing test cases. 
We employ the Pass$@k$ metric, where $k$ denotes the number of programs generated for each task.
A task is solved if at least one solution passes all unit tests, and we report the overall proportion of solved tasks. 
To reduce sampling variance, we generate $n \ge k$ solutions (for this study, $n = 20$ and $k=1, 5, 10$) for each task, count the number of correct solutions $c \le n$ that pass the unit tests, and calculate the unbiased estimator:
\begin{equation}
    \text{Pass}@k = \mathbb E\left[1-\frac{\binom{n-c}{k}}{\binom{n}{k}}\right]
\end{equation}


\section{Results and Analysis}
\label{sec_experiment_results}

\begin{figure}
    \centering
    \includegraphics[width=\linewidth]{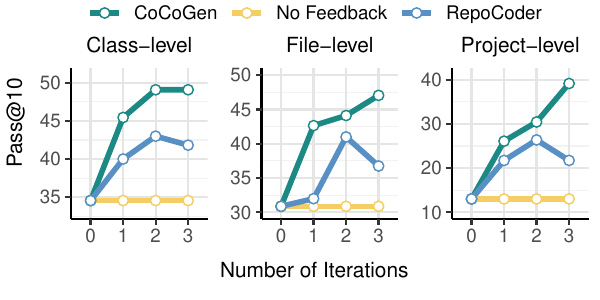}
    \vspace{-1em}
    \caption{Pass@10 of \tool, RepoCoder, and No Feedback baseline across three dependency levels}
    \vspace{-1em}
    \label{fig:iterative_analysis}
\end{figure}

\subsection{Overall Performance of \tool}
Table~\ref{table:main_result} 
shows the overall performance of \tool, assessed against various baselines, on the CoderEval~\cite{yu2023codereval} and the CrossCodeEval~\cite{ding2024crosscodeeval} dataset, respectively.
This table shows that the \tool can significantly outperform other baselines on the project-level code generation task.
This trend persists, with a few exceptions noted specifically in terms of Pass$@1$. 
We attribute such exceptions to variations in the generated solutions.
Selecting a significantly larger 
$n$ (\textit{e.g.}, $1000$), as discussed in
~\cite{li2022competition}, stabilizes the result and eliminates these expectations.
Moreover, it becomes evident that models incorporating contextual information, such as ReACC, RepoCoder, and \tool, exhibit a noteworthy performance superiority over the no context (\ie, direct) model, thereby affirming the practical value of the context in project-level code generation.

\subsection{Effectiveness of the Iterative Refinement}
Here, we investigate the effectiveness of iterative refinement in code generation with compiler feedback.
We conduct an ablation analysis on both RepoCoder and \tool, via removing or retaining the iterative refinement process.
Figure~\ref{fig:iterative_analysis} shows the performance of RepoCoder and \tool, with respect to varying iterations, on different data splits.
This figure clearly illustrates that as the number of iterations increases, the performance of \tool also exhibits a corresponding improvement.
The improvement substantiates the efficacy of our suggested iterative refinement process, demonstrating its ability to enhance the generated code through multiple iterations progressively.

\begin{table}
\caption{Pass rates of \tool with components ablated, based on GPT-3.5-Turbo model using the CoderEval-Python dataset}
\label{table:ablation_all}
\vspace{0em}
\begin{center}
\resizebox{1\columnwidth}{!}{
\begin{tabular}{llll}
\hline
Method                        & Pass@1 & Pass@5 & Pass@10 \\ \hline
\rowcolor[HTML]{EFEFEF} 
\tool          & 28.01  & \textbf{43.01}  & \textbf{46.58}   \\
- w/o CF and SQL (RepoCoder)  & \textbf{28.72}  & 34.44  & 36.30    \\
\rowcolor[HTML]{EFEFEF} 
- w/ CF, w/o SQL, w/o Semantic              & 25.69  & 37.50   & 41.78   \\
- w/ CF and SQL, w/o Semantic & 26.37  & 38.31  & 41.78   \\
\rowcolor[HTML]{EFEFEF} 
- w/ CF and Semantic, w/o SQL & 27.39  & 40.02  & 44.45   \\ \hline
\end{tabular}
}
\end{center}
\vspace{-1em}
\end{table}

\subsection{Ablation on Components}

To explore the impact of each component within \tool on overall performance, we remove various components from \tool, including the compiler, SQL retriever, and semantic retriever. Notably, the configuration that excludes the compiler and SQL, relying solely on semantic retrieval, is known as RepoCoder~\cite{zhang2023repocoder}.

Table~\ref{table:ablation_all}
demonstrates the pass rates of \tool across CoderEval-Python. 
The data indicates that incorporating compiler feedback does improve accuracy, although not markedly substantial. The reason is that highlighting compilation errors alone does not provide related context for resolving them, which is important in project-level coding problems.
Also, compared to RepoCoder—which relies solely on semantic retrieval, integrating compiler feedback with project-specific context results in a stable performance improvement. Results on each data split can be found in Appendix~\ref{apdx:more_ablation}.

Further analytical experiments, including the evaluation of benefits from solely compiler feedback, the performance of \tool on project-level code completion and function-level code generation, and the efficiency of static analysis and SQL queries, can be found in Appendix~\ref{apdx:more_experiments}.

\begin{figure}
    \centering
    \includegraphics[width=\linewidth]{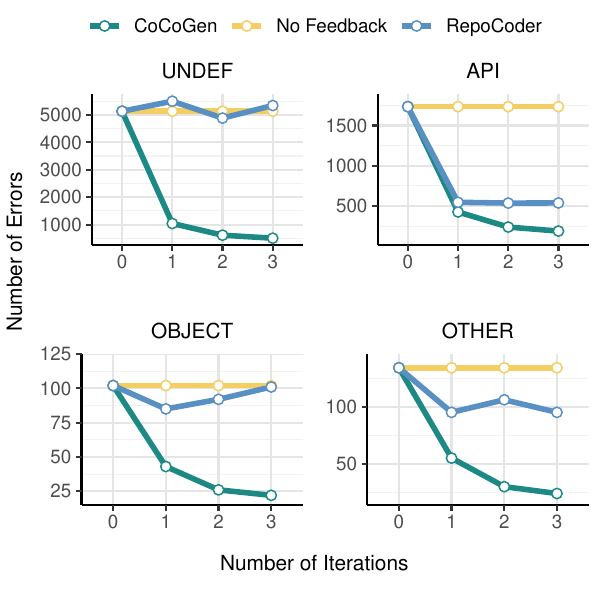}
    \vspace{-1em}
    \caption{Compilation errors  fixed per iteration of \tool, RepoCoder, and No Feedback baselines}
    \vspace{0em}
    \label{fig:error_statistic}
\end{figure}

\subsection{Error Analysis and Case Study}
We also perform an error analysis of the generated code in iterative generation.
We follow the categorization of errors defined in Section~\ref{sec:error_empirical}, examples of each error type can be found in Table~\ref{table:error_examples}.
Figure~\ref{fig:error_statistic} shows the distribution of errors resolved iteratively by our \tool and two baselines. 
From this figure, we can see that the errors of various types can be effectively resolved after a single iteration of refinement. For instance, \textit{UNDEF} errors are notably reduced from 5,133 to 1,042 after one iteration.
Additionally, it is observed that the RepoCoder baseline, which operates without compiler feedback, manages to rectify API and syntax errors, corroborating the findings in ~\citet{zhang2023repocoder}.
Nonetheless, RepoCoder proves ineffective against \textit{UNDEF} and \textit{OBJECT} errors, likely due to the model's lack of awareness regarding these errors in the absence of compiler feedback. 

\begin{figure}
    \centering
    \includegraphics[width=\linewidth]{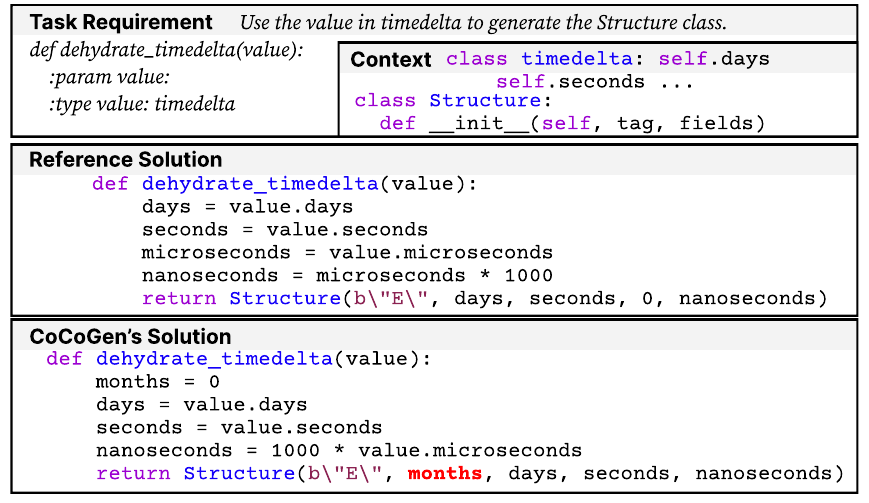}
    \vspace{-1em}
    \caption{An example of a runtime error in \tool's generated code for a CoderEval test case}
    \vspace{0em}
    \label{fig:case_study}
\end{figure}
To thoroughly evaluate \tool's effectiveness, we focus on scenarios where compilation is successful, but execution fails. 
In the case illustrated in Figure~\ref{fig:case_study}, \tool incorrectly excludes the \texttt{microseconds} field and erroneously adds a \texttt{month} field.
This error stems from ambiguously stated task requirements and the model's lack of familiarity with the \texttt{Structure} class's format, despite its definition being available, resulting in misinterpretation of the intended functionality. 
Further examples are detailed in Appendix~\ref{appendix:error_cases}.
The observations inspire us to integrate a comprehensive reference comprising documentation, web search results, and code execution log to provide explicit guidelines for code generation in our future work.

\section{Related Work}

\paragraph{LLM-based Code Generation.}
Automated code generation has a history spanning several decades, with initial endeavors utilizing rule-based systems~\cite{woods1973progress} and structured prediction~\cite{zelle1996learning,zettlemoyer2005learning}.
In recent years, the development of LLMs has led to the emergence of many prominent models in coding tasks. 
These include open-access models such as DeepSeek Coder~\cite{bi2024deepseek}, Code Llama~\cite{roziere2023code}, and StarCoder~\cite{li2023starcoder}, alongside commercial offerings like GPT-3.5~\cite{chatgpt} and GitHub Copilot~\cite{copilot}. These models and tools have demonstrated significant promise in enhancing code generation capabilities.

\paragraph{Project-level Code Generation.}
Generating accurate code within a project poses challenges due to the modular design of software engineering, which results in cross-file dependency patterns~\cite{parnas1972criteria}.
Early works augmented N-gram, RNN, and LSTM models with an additional cache model to track project-level changes~\cite{tu2014localness, hellendoorn2017deep}.
\citet{pashakhanloo2022codetrek} transformed projects into a relational database and proposes a graph walking method to traverse this database.
\citet{zan2022language} first used private API documentation to improve code generation, and \citet{zan2023private} investigated how various components of API documentation influence prediction accuracy.
\citet{shrivastava2023repository} proposed a code completion framework using a classifier to filter useful repository-level prompt proposals.
\citet{lu2022reacc, zhang2023repocoder} proposed to use single or multiple levels of retrieval-augmented generation mechanisms for code generation.
\citet{liao2023context} proposed A$^3$-CodGen, which utilizes local, global, and third-party-library information for better context retrieval.
\cite{yu2023codereval, liu2023repobench, zhou2022docprompting,ding2024crosscodeeval} proposed benchmarks and datasets for repository-level code generation tasks.

\paragraph{Post-processing of LLMs for Code Generation.}
To identify correct code generated by LLM, researchers employ post-processing techniques to further rank and filter the generated code.
\citet{inala2022fault} trained a fault-aware neural ranker that ranks multiple code samples based on compilation feedback.
AlphaCode~\cite{li2022competition} and~\citet{shi2022natural} employed filtering methods based on the execution feedback.
\citet{zhang2023coder} reranked LLM outputs based on back translation.
SelfEdit~\cite{zhang2023self} employed a generate-and-edit approach that utilizes execution results to improve the competitive programming task code quality.
\citet{chen2023teaching} proposed a rubber duck debugging mechanism without human feedback, which identifies mistakes in generated code by investigating the execution results, and explaining the generated code in natural language.

\section{Conclusion} 
In this paper, we show how to use project-specific context information, indexed structural and semantic, to fix compilation errors generated by compilers and improve the quality of code generated by LLMs.
Our experimental results show 
the increased prevalence of errors related to project contexts in project-level code generation compared to function-level code generation.
The presented \tool
can effectively fix the compilation errors by retrieving related context from the project, thus significantly improving the native LLM baselines on over 80\% relative pass rates in generating functions dependent on project-specific contexts.



\section*{Acknowledgements}
This work is supported by the Major Program (JD) of Hubei Province (Grant No. 2023BAA024), and the National Natural Science Foundation of China under grant No. 62102157. 
This work is also partially supported by Huawei.
Fangxin Lu is a visiting student at Huazhong University of Science and Technology, and she is from South-Central Minzu University for Nationalities.

\section*{Limitations}
In this paper, we utilize compilation information as a means to validate programs.
However, it is important to note that even programs that compile successfully can experience execution failures. Furthermore, the successful compilation of programs does not guarantee their safety for execution. As a result, while our findings indicate improvements in quality metrics through the correction of code to properly leverage context, it is imperative to undertake additional verification methods such as testing and manual review to ascertain the functional correctness of the generated code. The challenge of ensuring functional correctness of code encompasses various aspects, including compliance with task requirements, adherence to pre/post-conditions and security requirements, and preserving robustness in generating code. 
With many questions unanswered, we hope our study can promote a broader view of utilizing computational linguistic technologies in the realm of automated software engineering.

\section*{Ethics Statements}
We meticulously ensure that all code and models integrated into our research adhere to open-access policies as outlined by the Creative Commons license. The methodology ensures full compliance with copyright and intellectual property laws, thereby eliminating any potential for infringement or unauthorized use of protected materials. By exclusively utilizing resources that are freely available and legally distributable, we maintain the highest standards of ethical conduct in research. This approach fosters an environment of transparency and respect for the intellectual property rights of others. Our commitment to these principles ensures that our work advances the frontiers of knowledge in a manner that is both legally sound and ethically responsible.


\bibliography{ref}
\bibliographystyle{acl_natbib}

\appendix


\section{More Details on Investigated Large Language Models}
\label{apdx:models}

In this paper, we have selected GPT-3.5-Turbo and Code-Llama 13B for investigation.

\paragraph{GPT-3.5-Turbo~\cite{chatgpt}.}
It is a large-scale decoder-only model based on the Transformer architecture~\cite{vaswani2017attention}. It is pre-trained on a diverse array of data, encompassing both natural language and code, and can learn a specific task given an instruction and several demonstration examples.
We perform the inference by invocating the model through an online API. 

\paragraph{Code Llama~\cite{roziere2023code}.}
It is a family of LLMs for code based on Llama 2~\cite{touvron2023llama2} and further trained on 1TB code and 
natural language tokens. It can be used for code completion and generation following natural language instructions.
For Code-Llama 13B, we utilize its base variant, specifically designed for general code generation and understanding.\footnote{The model can be accessed via \url{https://huggingface.co/codellama/CodeLlama-13b-hf}.}

In the inference stage, 
we set the decoding temperature to $0.7$, and adopt the top-$k$ sampling strategy.
We implement the retrieval modules based on the \texttt{text-ada} model introduced by OpenAI~\cite{textada}, which is effective in both natural language search and code search.
The embedding dimension is 1,536 for the \texttt{text-ada} model.
We retrieve at most 5 entries for each query.
All the experiments in this paper are conducted on a Linux server with 128GB memory, with four 32GB Tesla V100 GPUs.

\section{More Details on the Investigated Dataset}
\label{apdx:dataset}
In this paper, we also use three recently proposed benchmarks evaluating the project-level code completion and function-level code generation ability. We provide the details of each benchmark.

\paragraph{CrossCodeEval~\cite{ding2024crosscodeeval}.} It is a project-level code completion benchmark that necessitates cross-file contextual understanding to complete the code accurately. CrossCodeEval is built on a diverse set of real-world, open-sourced, permissively-licensed repositories, including 471 repositories, 2,665 test cases across 1,368 files for Python language~\cite{ding2024crosscodeeval}.

\paragraph{HumanEval~\cite{chen2021evaluating}.} It is a function-level code generation benchmark that consists of 164 programming problems with corresponding Python solutions. These problems cover a range of difficulty levels and programming concepts. HumanEval is widely used in evaluating LMs for code generation tasks due to its focus on real-world coding scenarios and its comprehensive test cases.

\paragraph{MBPP~\cite{austin2021program}.} It is a function-level code generation benchmark consisting of 974 crowd-sourced Python programming problems. These problems are designed to be solvable by entry-level programmers and cover programming fundamentals and standard library functionalities. Each problem includes a task description, a code solution, and several automated test cases.

\section{More Details on the Evaluation Metrics}
\label{apdx:evaluation_metrics}
In this work, we use several evaluation metrics for method generation and line completion tasks.
This includes:
\paragraph{Exact Match.} 
This metric assesses whether each character of the model's predicted code exactly matches each character of the correct answer. All characters are the same yields $EM=1$, while any discrepancy results in $EM=0$. This is a strict all-or-nothing metric, where a single character difference results in a score of 0. The average value of EM on all test cases is reported.

\paragraph{Edit Similarity~\cite{levenshtein1966binary}.} This metric quantifies how dissimilar two strings are to one another. It is measured by counting the minimum number of operations required to transform one string into the other:
\begin{equation}
\resizebox{1\linewidth}{!}{$
\operatorname{lev}(a, b)= \begin{cases}|a| & \text { if }|b|=0, \\ |b| & \text { if }|a|=0, \\ \operatorname{lev}(\operatorname{tail}(a), \operatorname{tail}(b)) & \text { if } \operatorname{head}(a)=\operatorname{head}(b), \\ 1+\min \begin{cases}\operatorname{lev}(\operatorname{tail}(a), b) \\ \operatorname{lev}(a, \operatorname{tail}(b)) \\ \operatorname{lev}(\operatorname{tail}(a), \operatorname{tail}(b))\end{cases} & \text { otherwise }\end{cases}
$}
\end{equation}

\paragraph{F1 Score~\cite{schutze2008introduction}.}
This metric evaluates the balance between precision and recall for the retrieved identifiers. The F1 score is based on the number of identifiers that both the prediction and the ground truth share. Precision is defined as the ratio of the shared identifiers to the total number of identifiers in the prediction, while recall is the ratio of the shared identifiers to the total number of identifiers in the ground truth. The F1 score is then calculated as the harmonic mean of the recall rate and the precision rate:
\begin{equation}
    F_1=\frac{2}{\mathrm{recall}^{-1}+\mathrm{precision}^{-1}}
\end{equation}

\begin{table*}[t]
\caption{A complete list of frequently occurred errors reported by the compiler}
\label{table:detailed_error}
\vspace{-1em}
\begin{center}
\resizebox{1\linewidth}{!}{
\begin{tabular}{llp{2.5cm}l}
\hline
Error Category          & Error ID                                         & Corresponding Error Code      & Error Reason                                                                                                   \\ \hline
                        & \cellcolor[HTML]{EFEFEF}UNDEF-P                  & \cellcolor[HTML]{EFEFEF}E0401 & \cellcolor[HTML]{EFEFEF}Unable to import a Package.                                                            \\
                        & UNDEF-CM                                         & E1101                         & A Class is accessed for an unexistent Member.                                                                  \\
                        & \cellcolor[HTML]{EFEFEF}UNDEF-API                & \cellcolor[HTML]{EFEFEF}E0611 & \cellcolor[HTML]{EFEFEF}A function or API cannot be found in a module.                                         \\
\multirow{-4}{*}{UNDEF} & UNDEF-O                                          & E0602                         & An undefined variable or Object is accessed.                                                                   \\ \hline
                        & \cellcolor[HTML]{EFEFEF}API-TMA                  & \cellcolor[HTML]{EFEFEF}E1121 & \cellcolor[HTML]{EFEFEF}A function call passes Too Many positional Arguments.                                  \\
                        & API-IA                                           & E1120                         & A function call passes Insufficient Arguments.                                                                 \\
                        & \cellcolor[HTML]{EFEFEF}                         & \cellcolor[HTML]{EFEFEF}E1111 & \cellcolor[HTML]{EFEFEF}Assignment from the function that doesn't return anything.                             \\
\multirow{-4}{*}{API}   & \multirow{-2}{*}{\cellcolor[HTML]{EFEFEF}API-WA} & \cellcolor[HTML]{EFEFEF}E1123 & \cellcolor[HTML]{EFEFEF}A function call passes a keyword argument which has no corresponding formal parameter. \\ \hline
                        & OBJ-NI                                           & E1133                         & A Non-Iterable value is used in place where iterable is expected.                                              \\
                        & \cellcolor[HTML]{EFEFEF}OBJ-NC                   & \cellcolor[HTML]{EFEFEF}E1102 & \cellcolor[HTML]{EFEFEF}An object being called is a Non-Callable object.                                       \\
\multirow{-3}{*}{OBJ}   & OBJ-NS                                           & E1136                         & A subscripted value does Not support Subscription.                                                             \\ \hline
OTHER                   & \cellcolor[HTML]{EFEFEF}OTHER                    & \cellcolor[HTML]{EFEFEF}      & \cellcolor[HTML]{EFEFEF}Other errors reported by analyzer.                                                     \\ \hline
\end{tabular}

}
\end{center}
\vspace{-1em}
\end{table*}

\section{Errors Reported by the Compiler}
\label{appendix:detailed_error_types}
We utilize \texttt{pylint} for error checking; it is a static code analyzer designed to inspect code within a project without executing it. 
The analyzer accepts the entire file containing the solution along with associated source files as input, and then it extracts errors that pertain directly to the lines in the generated solution from the entirety of identified errors. 
\texttt{pylint} generates a range of diagnostic messages, encompassing errors, warnings, recommendations for adhering to language conventions, and suggestions for refactoring to adhere to best coding practices. However, our focus is solely on the errors.

Each error is identified by an error code and described with an error message. 
The error code is a numerical identifier that reflects the error's nature, while the error message provides a concise description of the error. 
\texttt{pylint} recognizes 133 distinct error types, and only 12 of these are reported frequently on the CoderEval-Python dataset.
We have categorized these errors into four groups based on their characteristics, detailed in Table~\ref{table:error_examples}. 
The \textit{FUNC} error category is also included in the error distribution analysis but is excluded from the compiler feedback pipeline. 
This exclusion is because such errors can only be identified with execution, a process that extends beyond the scope of static analysis and is unsuitable for the real-time generation and refinement pipeline. 
The specific types of each error and corresponding error codes are documented in Table~\ref{table:detailed_error}.

\begin{table}
\caption{Pass rates on CoderEval-Python of each approach with or without compiler feedback}
\label{table:compiler_feedback}
\vspace{-1em}
\begin{center}
\resizebox{1\columnwidth}{!}{
\begin{tabular}{lccc}
\hline
Method       & \multicolumn{1}{l}{Pass@1} & Pass@5 & Pass@10 \\ \hline
\rowcolor[HTML]{EFEFEF} 
Direct      & 20.65                      & 26.66  & 29.13   \\
Direct+CF   & 32.34                      & 38.43  & 40.43   \\ \hline
\rowcolor[HTML]{EFEFEF} 
ReACC        & 34.13                      & 41.44  & 43.48   \\
ReACC+CF     & 36.60                      & 44.05  & 46.52   \\ \hline
\rowcolor[HTML]{EFEFEF} 
RepoCoder    & 36.82                      & 40.73  & 42.17   \\
RepoCoder+CF & 38.00                      & 45.38  & 48.26   \\ \hline
\end{tabular}

}
\end{center}
\vspace{-1em}
\end{table}

\begin{table}[t]
\caption{
Pass rates of \tool on Class Runnable test cases
}
\label{table:ablation1}
\vspace{-1em}
\begin{center}
\resizebox{1\linewidth}{!}{
\begin{tabular}{llll}
\hline
Data Split     & \multicolumn{3}{c}{\textbf{Class Runnable}}      \\ \hline
Method         & Pass@1         & Pass@5         & Pass@10        \\ \hline
\rowcolor[HTML]{EFEFEF} 
\tool          & 28.00          & 44.92          & \textbf{49.09} \\
- w/o CF and SQL (RepoCoder) & \textbf{35.45} & 40.46 & 41.82 \\
\rowcolor[HTML]{EFEFEF} 
- w/ CF, w/o SQL, w/o Semantic          & 30.00          & 42.58          & 45.45          \\
- w/ CF and SQL, w/o Semantic      & 30.36          & 44.61          & 49.09          \\
\rowcolor[HTML]{EFEFEF} 
- w/ CF and Semantic, w/o SQL & 31.45 & \textbf{45.00} & 47.27          \\ \hline
\end{tabular}
}
\end{center}
\vspace{-1em}
\end{table}

\begin{table}[t]
\caption{
Pass rates of \tool on File Runnable test cases
}
\label{table:ablation2}
\vspace{-1em}
\begin{center}
\resizebox{1\linewidth}{!}{
\begin{tabular}{llll}
\hline
Data Split     & \multicolumn{3}{c}{\textbf{File Runnable}}       \\ \hline
Method         & Pass@1         & Pass@5         & Pass@10        \\ \hline
\rowcolor[HTML]{EFEFEF} 
\tool          & \textbf{30.29} & \textbf{43.58} & \textbf{47.06} \\
- w/o CF and SQL (RepoCoder) & 29.41 & 34.61 & 36.76 \\
\rowcolor[HTML]{EFEFEF} 
- w/ CF, w/o SQL, w/o Semantic          & 26.03          & 37.41          & 42.65          \\
- w/ CF and SQL, w/o Semantic      & 26.76          & 37.68          & 39.71          \\
\rowcolor[HTML]{EFEFEF} 
- w/ CF and Semantic, w/o SQL & 27.35          & 39.90          & 45.45          \\ \hline
\end{tabular}
}
\end{center}
\vspace{-1em}
\end{table}

\begin{table}[t]
\caption{
Pass rates of \tool on Project Runnable test cases
}
\label{table:ablation3}
\vspace{-1em}
\begin{center}
\resizebox{1\linewidth}{!}{
\begin{tabular}{llll}
\hline
Data Split     & \multicolumn{3}{c}{\textbf{Project Runnable}}    \\ \hline
Method         & Pass@1         & Pass@5         & Pass@10        \\ \hline
\rowcolor[HTML]{EFEFEF} 
\tool          & \textbf{21.30} & \textbf{36.73} & \textbf{39.13} \\
- w/o CF and SQL (RepoCoder) & 16.96 & 19.57 & 21.74 \\
\rowcolor[HTML]{EFEFEF} 
- w/ CF, w/o SQL, w/o Semantic          & 14.35          & 25.62          & 30.43          \\
- w/ CF and SQL, w/o Semantic      & 15.65          & 25.12          & 30.43          \\
\rowcolor[HTML]{EFEFEF} 
- w/ CF and Semantic, w/o SQL & 17.83          & 28.50          & 34.78          \\ \hline
\end{tabular}
}
\end{center}
\vspace{-1em}
\end{table}

\begin{figure}
    \centering
    \includegraphics[width=\linewidth]{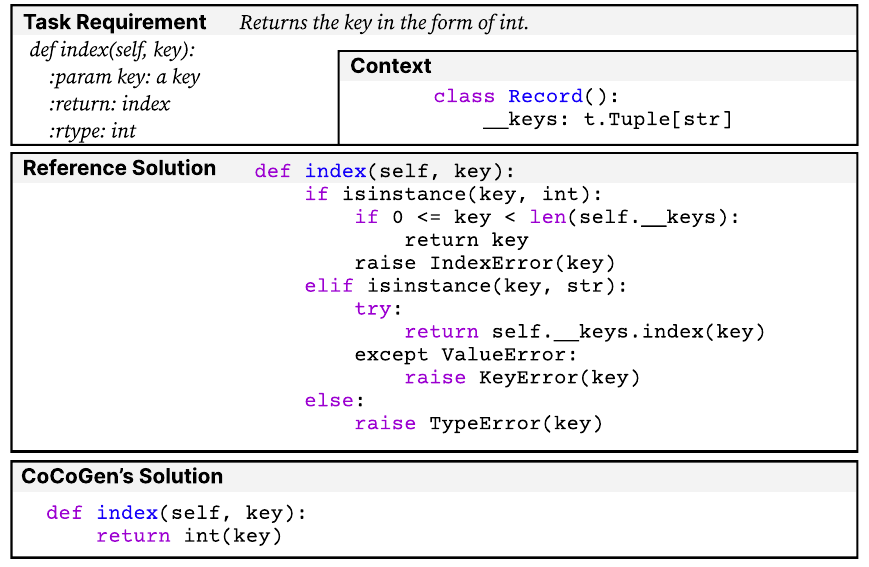}
    \vspace{-2em}
    \caption{An error case of degenrate solution}
    \vspace{-1em}
    \label{fig:extra_case_study_1}
\end{figure}

\begin{figure}
    \centering
    \includegraphics[width=\linewidth]{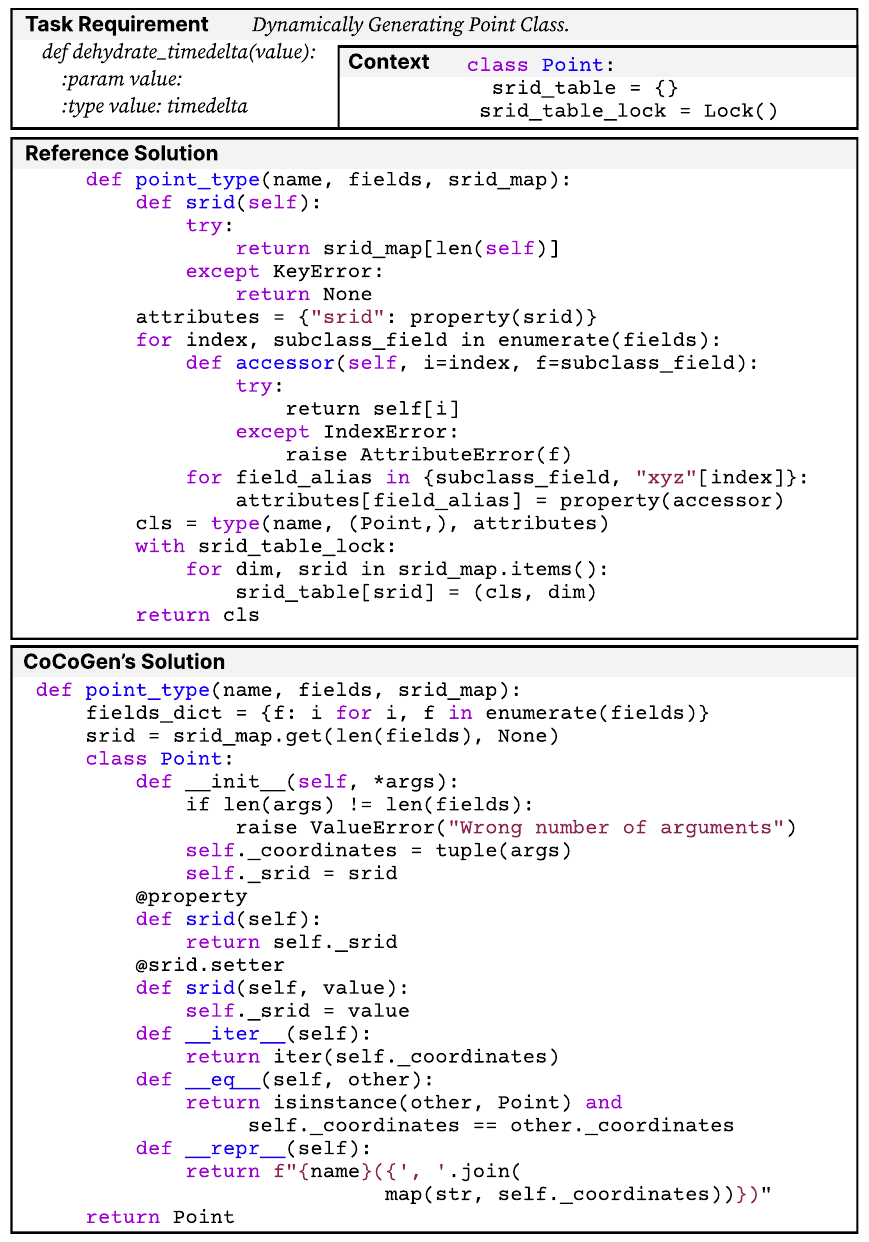}
    \vspace{-2em}
    \caption{An error case of misinterpreting task requirements}
    \vspace{-1.5em}
    \label{fig:extra_case_study_2}
\end{figure}

\section{More Details About the Error Cases}
\label{appendix:error_cases}
We delve into additional cases where \tool struggles to generate viable solutions, elucidating the underlying causes to foster further research. 
We identify two primary factors leading to execution errors despite successful compilation. The first factor is the occurrence of degenerate solutions, such as overly simplistic or redundant code. This phenomenon is discovered and explored in~\cite{zhang2023coder}.
Figure~\ref{fig:extra_case_study_1} exemplifies this issue, showcasing a solution that omits necessary validity checks and error handling, neither of which are specified in the task requirements, nor produced by \tool's outputs. Our analysis reveals that degenerate solutions frequently pass compilation, rendering compiler-based verification ineffective.

Another prevalent mistake involves misinterpreting task requirements, resulting in solutions that lack logical coherence. Figure~\ref{fig:extra_case_study_2} depicts an instance where the assignment is to leverage the pre-existing \texttt{Point} class; instead, the LM disregards this specification and redundantly recreates the class. This underscores the challenge LLMs face in accurately comprehending prompts and generating appropriate solutions.

\section{More Analytic Experiments}
\label{apdx:more_experiments}

\subsection{Usefulness of Compiler Feedback}
Here, we examine the usefulness of compiler feedback by integrating it into three baseline models: Vanilla, ReACC, and RepoCoder, each considered separately.
We conduct experiments using the GPT-3.5-Turbo and subsequently reported the average score across the entire CoderEval-Python dataset consisting of six levels of context dependencies, as shown in Table~\ref{table:compiler_feedback}.
The table provides clear evidence that incorporating compiler feedback yields a notable enhancement in model performance for code generation. Specifically, a comparison between RepoCoder with and without compiler feedback reveals a substantial increase in Pass$@1$ from 36.82 to 38.00.

\subsection{More Results on the Ablation Study}
\label{apdx:more_ablation}
To further examine the performance of each component of \tool across different dependency levels of test cases, we evaluate \tool at various dependency levels from the CoderEval-Python test suite. 
The results are presented in Table~\ref{table:ablation1}, Table~\ref{table:ablation2}, and Table~\ref{table:ablation3}.
The results show that utilizing compiler feedback and structural queries consistently demonstrates performance improvements across most scenarios. Furthermore, it achieves better performance when stronger cross-file dependencies are present (\textit{i.e.}, at the file-level and project-level tasks), indicating that \tool can accurately capture the context across the project.

\subsection{Varying the Number of Iterations}
\begin{table}[t]
\caption{
Pass rates of \tool on varying iterations
}
\label{table:iteration}
\vspace{-1em}
\begin{center}
\resizebox{1\linewidth}{!}{
\begin{tabular}{llll}
\hline
\textbf{Iteration} & \textbf{Class Level} & \textbf{File Level} & \textbf{Project Level} \\ \hline
\rowcolor[HTML]{EFEFEF} 
$i=0$         & 34.55               & 30.88              & 13.04                 \\
$i=3$         & \textbf{49.09}               & \textbf{47.06}              & \textbf{39.13}                 \\
\rowcolor[HTML]{EFEFEF} 
$i=10$        & 47.27               & 45.59              & \textbf{39.13}                 \\ \hline
\end{tabular}

}
\end{center}
\vspace{-1em}
\end{table}

To investigate whether iteration could continuously improve performance, we increase the number of iterations to ten. As shown in Table~\ref{table:iteration}, with more iterations, the performance of \tool reaches a plateau. This indicates that there are still errors that cannot got repaired by \tool, leaving for further research.

\subsection{\tool on Project-level Code Completion}
We also evaluate \tool on project-level code completion tasks. Although \tool is originally designed for code generation tasks, the compiler feedback may also benefit this code completion task.
The results are presented in Table~\ref{table:crosscodeeval}.
From the table, we can see that the compiler feedback does improve performance, although not as significantly as it does in the code generation task.

\begin{table}
\caption{C-EM, C-ES, I-EM, and I-F1 scores based on the GPT-3.5-Turbo on the CrossCodeEval dataset}
\label{table:crosscodeeval}
\vspace{-1em}
\begin{center}
\resizebox{1\columnwidth}{!}{
\begin{tabular}{llllll}
\hline
Category  & \multicolumn{2}{l}{Code Match} &  & \multicolumn{2}{l}{Identifier Match} \\ \cline{1-3} \cline{5-6} 
Method    & C-EM          & C-ES           &  & I-EM              & I-F1             \\ \hline
\rowcolor[HTML]{EFEFEF} 
Direct    & 1.91          & 50.51          &  & 3.60              & 32.27            \\
ReAcc     & 5.48          & 54.03          &  & 10.02             & 38.49            \\
\rowcolor[HTML]{EFEFEF} 
RepoCoder & 8.52          & 55.05          &  & 13.02             & 41.05            \\
CoCoGen   & \textbf{9.08} & \textbf{55.31} &  & \textbf{14.11}    & \textbf{42.34}   \\ \hline
\end{tabular}
}
\end{center}
\vspace{-1em}
\end{table}

\subsection{\tool on Function-level Code Generation}
\begin{table}[t]
\caption{
The performance of CoCoGen across varying iterations on the HumanEval benchmark (CE: Compilation Errors, CE\%: their proportion among all generated solutions)
}
\label{table:humaneval}
\vspace{-1em}
\begin{center}
\resizebox{1\linewidth}{!}{
\begin{tabular}{llllll}
\hline
       & \textbf{Pass@1} & \textbf{Pass@5} & \textbf{Pass@10} & \textbf{CE} & \textbf{CE(\%)} \\ \hline
\rowcolor[HTML]{EFEFEF} 
Direct & 61.52           & 80.56           & 83.15            & 40          & 2.38\%          \\
CF $i=1$ & 71.46           & 82.46           & 85.20             & 7           & 0.42\%          \\
\rowcolor[HTML]{EFEFEF} 
CF $i=2$ & 71.40            & 82.56           & 85.98            & 3           & 0.18\%          \\
CF $i=3$ & 71.65           & 82.49           & 85.36            & 3           & 0.18\%          \\ \hline
\end{tabular}
}
\end{center}
\vspace{-1em}
\end{table}

\begin{table}[t]
\caption{
The performance of CoCoGen across varying iterations on the MBPP benchmark (CE: Compilation Errors, CE\%: their proportion among all generated solutions)
}
\label{table:mbpp}
\vspace{-1em}
\begin{center}
\resizebox{1\linewidth}{!}{
\begin{tabular}{llllll}
\hline
       & \textbf{Pass@1} & \textbf{Pass@5} & \textbf{Pass@10} & \textbf{CE} & \textbf{CE(\%)} \\ \hline
\rowcolor[HTML]{EFEFEF} 
Direct & 49.71           & 58.50           & 60.81            & 185         & 1.90\%          \\
CF $i=1$ & 52.95           & 59.58           & 61.85            & 11          & 0.11\%          \\
\rowcolor[HTML]{EFEFEF} 
CF $i=2$ & 53.06           & 59.87           & 62.01            & 9           & 0.09\%          \\
CF $i=3$ & 52.79           & 59.80           & 62.09            & 7           & 0.07\%          \\ \hline
\end{tabular}

}
\end{center}
\vspace{-1em}
\end{table}

To assess the efficacy of \tool in function-level code generation, we evaluate it on two commonly used datasets: HumanEval~\cite{chen2021evaluating} and MBPP~\cite{austin2021program}. 
We retain the compiler feedback model in \tool, and conduct tests in a 0-shot scenario, generating 10 code samples for each of the 163 test cases for HumanEval and 974 cases for MBPP, totaling 11,370 samples. The generation and refinement loop iterates 3 times, as shown in Table~\ref{table:humaneval} and Table~\ref{table:mbpp}.

The data presented in the table illustrates that compilation feedback improves the accuracy of the generated code, although its efficacy is somewhat limited. We manually review 7 compilation errors not resolved in HumanEval after the first iteration, and find that 6 out of 7 instances are caused by the LLM mistakenly generating code segments enclosed in Markdown code block markers (\texttt{`\unskip `\unskip `}), resulting in compilation failures. 
Additionally, another instance of compilation error is identified in the following generated solution:

\begin{tcolorbox}[colback=white, colframe=black, sharp corners, left=1pt,right=1pt, top=1pt, bottom=1pt,boxrule=1pt]
\texttt{
def is\_simple\_power(x, n):
return x > 0 and (x == 1 or (n != 1 and x == n**int(round(math.log(x, n))))
}
\end{tcolorbox}

This code snippet contains five left parentheses and four right parentheses in the \texttt{return} statement, causing the syntax error. Interestingly, despite the compiler has indicated this syntax error, it is not rectified over three iterations.

In MBPP, we manually observe the 7 errors not fixed after the last iteration, and identify that two test tasks are responsible for all 7 errors: one named "sum" contributed to six errors, and another called "month\_season" resulted in one error. We detail these error cases as follows:

\begin{tcolorbox}[colback=white, colframe=black, sharp corners, left=1pt,right=1pt, top=1pt, bottom=1pt,boxrule=1pt]
\texttt{
def month\_season(month, day):
    seasons = {
        "spring": [(3, 20), (6, 20)],
        "summer": [(6, 21), (9, 22)],
        "autumn": [(9, 23), (12, 20)],
        "winter": [(12, 21), (3, 19)]
    }
    for season, (start, end) in seasons.items():
        if (month == start[0] and day >= start[1]) or (month == end[0] and day <= end[1]):
            return season
    return "Invalid input"
}
\end{tcolorbox}
The error arises because \texttt{'start'} and \texttt{'end'} are not included in \texttt{seasons.items}, which appears strange. We speculate that this issue stems from the presence of functions with the same name within the training set.

\begin{tcolorbox}[colback=white, colframe=black, sharp corners, left=1pt,right=1pt, top=1pt, bottom=1pt,boxrule=1pt]
\texttt{
def sum(a, b):
    common\_divisors = [i for i in range(1, min(a, b) + 1) if a \% i == 0 and b \% i == 0]
    return sum(common\_divisors) if common\_divisors else 0
}
\end{tcolorbox}
The root cause of this error lies in the model's confusion when a user-defined function name coincides with that of a system library function \texttt{sum}, representing a category of potential issues.

\begin{table*}[t]
\caption{Tables pre-computed by \tool for error correction}
\label{table:sql_tables}
\vspace{-1em}
\begin{center}
\resizebox{1\linewidth}{!}{
\begin{tabular}{lll}
\hline
\textbf{Table Name} & \textbf{Description}                                         & \textbf{Element Example}                                                    \\ \hline
\rowcolor[HTML]{EFEFEF} 
\textbf{M}          & Stores the a module and its hierarchy in project.            & tests.unit.async\_.work.\_\_init\_\_                                        \\
\textbf{M\_C}       & Stores a module and a class inside the module                & Module neo4j.\_codec.packstream.v1.\_\_init\_\_, Class PackableBuffer       \\
\rowcolor[HTML]{EFEFEF} 
\textbf{M\_C\_CF}   & Stores a class, its parent module, and its member functions. & Module neo4j.time.\_\_init\_\_, Class Clock, Function local\_offset         \\
\textbf{M\_C\_V}             & Stores a class variable, its parent class and module.        & Module neo4j.\_sync.io.tmphhoug1of, Class Bolt, Variable is\_reset          \\
\rowcolor[HTML]{EFEFEF} 
\textbf{M\_GF}               & Store a global function and its parent module.               & Module neo4j.time.\_arithmetic, Function nano\_add                          \\
\textbf{M\_GV}               & Stores a global variable and its parent module.              & neo4j.\_\_init\_\_, Global Variable TRUST\_SYSTEM\_CA\_SIGNED\_CERTIFICATES \\ \hline
\end{tabular}

}
\end{center}
\vspace{0em}
\end{table*}

\begin{table*}[t]
\caption{A complete list of demonstration examples prompted to the language model}
\label{table:sql_example}
\vspace{-1em}
\begin{center}
\resizebox{1\linewidth}{!}{
\begin{tabular}{l|p{5cm}p{5cm}p{6cm}}
\hline
\textbf{Error Type} & \textbf{Example Error Message}                                      & \textbf{Action}                                               & \textbf{Example Structural Query}                                                                                                                                                                          \\ \hline
\rowcolor[HTML]{EFEFEF} 
UNDEF-P    & Unable to import 'keys'                                    & Confine the search scope in all modules              & \texttt{from Module m  where m.inSource()  and v.getScope() = m  select m}                                                                                                                        \\
UNDEF-CM   & Instance of 'RootLogger' has no 'loggerDict' member        & Confine the search scope in all members in the class & \texttt{from Module m, Class c, Function cf where m.inSource() and m.contains(c) and c.contains(cf) and cf.getScope() = c and c.getName = 'RootLogger' and not cf.isInitMethod() select m, c, cf} \\
\rowcolor[HTML]{EFEFEF} 
UNDEF-API  & No name 'AsyncBolt5x0' in module 'neo4j.\_sync.io.\_bolt5' & Confine the search scope in all names in the module  & \texttt{from Module m, Variable v where m.inSource()  and v.getScope() = m  and m.getName() = 'neo4j.\_sync.io.\_bolt5' select m, v.getDefinition()}                                              \\
API        & No value for argument 'xmls' in function call 'dumpXML'    & Return the information of the function               & \texttt{from Module m, Function f    where m.inSource()  and m.contains(f)    and f.getName() = 'dumpXML'  select m, f}                                                                           \\ \hline
\end{tabular}

}
\end{center}
\vspace{0em}
\end{table*}

\subsection{Efficiency of Static Analysis}

Utilizing static analysis tools to inspect code typically results in increased latency. To investigate whether the latency impacts the usability of \tool, we measure its latency at each test case. 
We have observed that in experiments, the speed of static analysis and structural query is relatively fast. It is because the semantic checker (i.e., \texttt{pylint}), only analyzes the current file and its dependent files. The average latency is recorded at 1.27 seconds, with a minimum of 0.359 seconds and a maximum of 6.984 seconds. We deem this latency to be acceptable.

\section{Technical Details of Structural Query System}
\label{apdx:sql_system}
\subsection{The Design Principle}
In \tool, we employ the \textit{Structural Query Language} (SQL) to perform structured queries on code repositories.
The SQL syntax used here is CodeQL, a specialized version designed for software repository mining, capable of conducting complex control-flow and data-flow analyses on a set of code files. 

We extracted 11 error codes from the compiler (shown in Table~\ref{table:detailed_error}), each corresponding to one of four error categories as in Section~\ref{sec:error_empirical}. 
For efficiency, two authors manually inspected the most frequently occurring error codes, precomputed six data tables from the project graph, and hard-coded the structural context retrieval procedure for the following error codes: \texttt{E0001} (syntax error), \texttt{E0602} (undefined-variable), \texttt{E1101} (no-member), \texttt{E0213} (no-self-argument), and \texttt{E0102} (function redefined).
The tables to retrieve structural project context for these frequently occurred errors are presented in Table~\ref{table:sql_tables}.

Whenever the compiler reports an error, if it is a frequently occurring error, its retrieval is done from these precomputed data tables. Otherwise, the \tool invokes the LLM to generate an SQL query statement, which is then executed on the project graph.

In the CoderEval-Python benchmark, with iteration rounds set to 3, related project contexts of 93.2\% of errors are successfully retrieved. For the remaining 6.8\% of cases, there are two reasons for failure: first, the LLM does not follow the demonstrated CodeQL examples and generates queries that are syntactically or semantically incorrect, thus rejected by the CodeQL query system; second, the query does not return anything because the specified conditions in the query are not satisfied in project context.

\onecolumn

\begin{multicols}{2}
\subsection{Demonstration Examples of Structural Queries}
\label{appendix:example_queries}
We present a demonstration example in Section~\ref{sec:retrieval_augmented} to compose the structural query.
The example focuses on identifying and addressing instances of missing or incorrectly utilized context entries. We detail several example error messages along with their corresponding structural queries.
A comprehensive list of these examples can be found in Table~\ref{table:sql_example}.
The four queries shown in the table are written manually by one of the authors, and designed to handle four representative types of compilation errors.
These sampled error messages are reported by compilers when generating solutions without project context.
The specific four messages are chosen randomly from the compilation log.

\section{Algorithm for Project Database Construction}
We present the comprehensive algorithm for constructing the project database and generating code with \tool.
The algorithm for building the project database is detailed in Algorithm~\ref{alg:database}. It involves identifying source files in the project by extracting all files that end with a \texttt{.py} extension.
To parse Python source files, we employ the \texttt{tree-sitter-python} parser for generating abstract syntax trees and \texttt{codeql-python} to extract the property of a context entry node.
To encode passages to vectors, we utilize \texttt{text-embedding-ada-002}, a text embedding model provided by OpenAI and accessible via online APIs~\cite{textada}.

\end{multicols}

\begin{algorithm*}
\caption{Project Database Construction}
\label{alg:database}
\begin{algorithmic}[1]
\REQUIRE $\textsc{SourceFileSet}$: A set of project source files
\REQUIRE $\textsc{Parser}$: A parser for source files
\REQUIRE $\textsc{Encoder}$: A passage encoder transforms text to numerical vector
\ENSURE $databaseEntries$: Entries in the project database

\STATE $databaseEntries \leftarrow \emptyset$
\FOR{each $sourceFile$ in $\textsc{SourceFileSet}$}
    \STATE $nodesForVisit \leftarrow \langle \rangle$
    \STATE $propertyPrefixSeq \leftarrow \langle \rangle$
    \STATE $astFile \leftarrow \textsc{Parser}(sourceFile)$
    \STATE $nodesForVisit.\textsc{add}(astFile.rootNode)$
    \WHILE{$nodesForVisit$ is not empty}
        \STATE $currentNode \leftarrow nodesForVisit.\textsc{pop}()$
        \IF{$currentNode$ is \textsc{PrefixMark}}
            \STATE $propertyPrefixSeq.\textsc{pop}()$
        \ENDIF
        \IF{$currentNode.type$ is in $\left[\textsc{VariableType, FunctionType, ClassType}\right]$}
            \STATE $nodeProperty \leftarrow \textsc{GetProperties}(currentNode)$
            \STATE $nodeSchema \leftarrow \langle propertyPrefixSeq, currentNode, nodeProperty \rangle$
            \STATE $nodeEmbedding \leftarrow \textsc{Encoder}(nodeSchema)$
            \STATE $databaseEntries.\textsc{add}(\left[nodeSchema, nodeEmbedding\right])$
            \STATE $propertyPrefixSeq.\textsc{push}(currentNode)$
            \STATE $nodesForVisit.\textsc{push}(\textsc{PrefixMark})$
        \ENDIF
        \FOR{$childNode$ in $currentNode.\textsc{childs}()$}
            \STATE $nodesForVisit.\textsc{push}(childNode)$
        \ENDFOR
    \ENDWHILE
\ENDFOR
\end{algorithmic}
\end{algorithm*}

\end{document}